\documentclass[10pt,conference,a4paper]{IEEEtran}
\IEEEoverridecommandlockouts
\usepackage{cite}
\usepackage{amsmath,amssymb,amsfonts}
\usepackage{algorithmic}
\usepackage{textcomp}
\usepackage{xcolor}
\usepackage[T1]{fontenc}

\usepackage[breaklinks=true,bookmarks=false]{hyperref}
\hypersetup{
  colorlinks   = true, 
  urlcolor     = blue, 
  linkcolor    = blue, 
  citecolor   = red 
}
\usepackage{booktabs}
\usepackage{multirow}
\usepackage{float}
\usepackage{tabularx}
\usepackage[hypcap=false]{caption}
\usepackage{graphicx}
\usepackage{graphbox}
\usepackage{cleveref}
\usepackage[section]{placeins}
\usepackage{etoolbox,lipsum}
\usepackage{afterpage}
\usepackage{subcaption}
\usepackage[autostyle, english = american]{csquotes}
\MakeOuterQuote{"}
\usepackage{capt-of,etoolbox}
\usepackage[section]{placeins}
\usepackage{etoolbox,lipsum}
\usepackage{multirow, makecell}
\usepackage{multirow}
\usepackage{xcolor,colortbl}
\usepackage{soul}
\usepackage{color}
\newcommand{\mypar}[1]{\vspace{0.3cm}\noindent\textbf{#1}}
\usepackage{expl3}
\usepackage{import}
\usepackage{svg}
\usepackage{pdfpages}
\usepackage{roboto}
\usepackage{xspace}
\usepackage{xfrac}
\ExplSyntaxOn
\newcommand\latinabbrev[1]{
  \peek_meaning:NTF . {
    #1\@}%
  { \peek_catcode:NTF a {
      #1.\@ }%
    {#1.\@}}}
\ExplSyntaxOff
\def\eg{\latinabbrev{e.g}}
\def\etal{\latinabbrev{et al}}

\def\ie{\latinabbrev{i.e}}

\newcommand\dataset{\emph{pic2kcal}\xspace}

\usepackage{balance}

\begin{document}
\title{Multi-Task Learning for Calorie Prediction on a \\ Novel Large-Scale Recipe Dataset \\ Enriched with Nutritional Information}

\author{
Robin Ruede\footnotemark$^1$  \quad Verena Heusser$^1$ \quad Lukas Frank$^1$ \quad Alina Roitberg$^2$  \quad  Monica Haurilet$^2$ \quad Rainer Stiefelhagen$^2$
\\Institute for Anthropomatics and Robotics, Karlsruhe Institute of Technology, Germany
\\  $^1${\tt\small \{firstname.lastname\}@student.kit.edu}, $^2${\tt\small \{firstname.lastname\}@kit.edu}
}
\maketitle

\begin{abstract}
A rapidly growing amount of content posted online, such as food recipes, opens doors to new exciting applications at the intersection of vision and language. 
In this work, we aim to estimate the calorie amount of a meal directly from an image by learning from recipes people have published on the Internet, thus skipping time-consuming manual data annotation.
Since there are few large-scale publicly available datasets captured in unconstrained environments, we propose the \dataset benchmark comprising 308\,000 images from over 70\,000 recipes including photographs, ingredients, and instructions.
To obtain nutritional information of the ingredients and automatically determine the ground-truth calorie value, we match the items in the recipes with structured information from a food item database.

We evaluate various neural networks for regression of the calorie quantity and extend them with the multi-task paradigm.
Our learning procedure combines the calorie estimation with prediction of proteins, carbohydrates, and fat amounts as well as a multi-label ingredient classification. 
Our experiments demonstrate clear benefits of multi-task learning for calorie estimation, surpassing the single-task calorie regression by \(\mathbf{9.9}\textbf{\%}\).
To encourage further research on this task, we make the code for generating the dataset and the models publicly available. 

\end{abstract}


\section{Introduction}

\begin{figure}[t]
  \centering
  \sffamily\robotocondensed\footnotesize 

\begin{tabular}{cc}

        \includegraphics[align=c,width=0.5\linewidth]{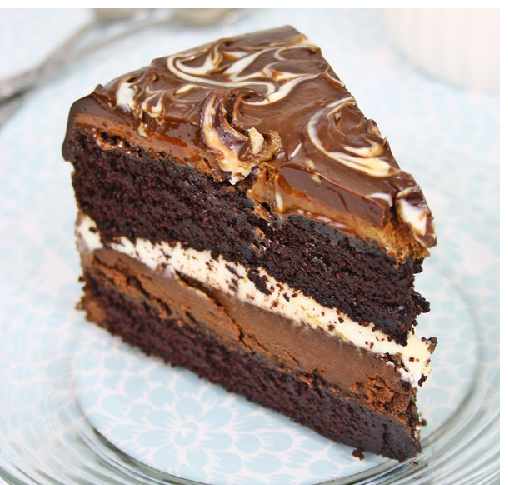} &    
        \begin{tabularx}{0.43\linewidth}{p{0.1\linewidth}>{\raggedleft}p{0.1\linewidth}>{\raggedleft\arraybackslash}p{0.1\linewidth}}
        \multicolumn{3}{l}{Nutrition Facts (per 100 g)} \\
        \toprule
        & Pred & True \\
        \midrule
        Calories & 183 kcal & 198 kcal \\
        \midrule
        Fat & 9 g & 9 g \\
        \midrule
        Carb & 17 g & 24 g \\
        \midrule
        Protein & 7 g & 4 g \\
        \midrule
        \multicolumn{3}{p{0.37\linewidth}}{\raggedright Ingredients (pred): Flour, Butter, Milk} \\
        \midrule
        \multicolumn{3}{p{0.37\linewidth}}{\raggedright Ingredients (true): Eggs, Flour, Vanilla Sugar} \\
        \bottomrule
        \end{tabularx} \\[15ex]

\end{tabular}

  \caption{Example image with accompanying nutritional values. The table shows the macronutrients and ingredients predicted by our model (pred) in comparison with the labels (true). Note that in this example we only consider the top-100 ingredients.
  }
  \label{fig:intro}
\end{figure}

A question that many of us have asked ourselves is how healthy our diet is and what we should change for a more balanced lifestyle.
Even though various diets and nutritional courses inform people how to live healthier, many still struggle to apply these rules in practice. 
Tracking the exact amount and quality (\eg, calories and macronutrient composition) of food eaten is important to successfully follow most diets. 
However, this can be particularly difficult, for instance when eating in a restaurant or canteen. 
Even when cooking ourselves, calculating these factors manually takes a lot of time and effort, leading to quick motivation loss and non-optimal results for many people. 
Guessing these values directly is also difficult as some studies show that most people are not able to correctly pinpoint the caloric value of a meal, mostly underestimating it~\cite{zhou2018calorie}. \looseness=-1

In this work, we propose an approach for predicting the calorie content of one portion \emph{directly}, in an end-to-end fashion from a single image of a meal.
Since there is little work on calorie estimation and most of them only deal with images captured in constrained environments~\cite{subhi2019vision}, we collect a large-scale dataset of pictures of meals captured in the ``wild''.
Our \dataset dataset comprises 308\,000 images of 70\,000 different recipes ranging from simple salads to cakes (\cref{fig:intro}), pizzas, and soups.
To collect our dataset and generate precise ground truth estimates, we propose a procedure based on the ingredients found in the recipe, instead of relying on user-given calorie estimates which are sparse and often incorrect. 
The ingredients and their mass given in the semi-structured section of the original recipe are each mapped to fully structured data such as calorie, macronutrient, and micronutrient amounts with the help of semantic embeddings created via free-text language models. 
These values are then aggregated to obtain the corresponding information for the finished dish.

We leverage our proposed dataset and implement different neural architectures for the calorie estimation task.
To that end, we pose the calorie estimation task as a regression problem and evaluate convolutional neural networks (CNNs) previously used for image recognition by minimizing the L1 loss. 
We improve the performance of these conventional CNNs for image recognition and propose a network trained in a multi-task setting where additional nutritional information and ingredients are estimated.
To foster further research in calorie estimation from unconstrained images, as well as for addressing other tasks concerning images of meals with structured recipe information, we make our code publicly available to the community\footnote{Code available at \url{https://github.com/phiresky/pic2kcal/}}\!.\looseness=-1


Overall,  the  contributions of this work are three-fold:\\
(1) We introduce a framework for retrieving nutritional information of recipes by matching ingredients and their mass to a nutrient database using phrase embeddings.\\
(2) We build a large-scale recipe dataset consisting of naturally collected dish images and their recipes collected from a German cooking website.\\
(3) We propose a \emph{multi-task} end-to-end approach for calorie and nutrient estimation based on a single image of the dish. 
\looseness=-1


\section{Related Work}
\label{sec:related_work}

\mypar{Calorie Estimation. }
In existing proprietary products, calorie tracking is mostly carried out manually by looking up specific ingredients and amounts for each eaten meal.
There exist a large number of automated dietary tracking approaches using mobile applications such as~\cite{noauthor_calorie_nodate}.
However, these require a lot of user intervention and a high amount of manual input.
For instance, the users are required to specify portion sizes and sometimes even ingredients manually, making this approach time-intensive and error prone.

In this work, we address the task of vision-based food analysis, which aims to predict nutritional information based on images of the dish.
The majority of vision-based works for food analysis (overview of methods and datasets in \cite{subhi2019vision}) consist of a multi-stage procedure~\cite{myers,chokr,pouladzadehUsingGraphCut2014,zhuIMAGEANALYSISSYSTEM2010}.
In these approaches, the image is segmented pixel-wise into food- and non-food components, which is followed by the classification of the image into a fixed set of categories, such as cuisine or food type.
The next stage consists of volume and weight estimation, and a nutrient information prediction step.
The nutritional information is predicted based on extracted features from pre-trained CNNs restricted by the segments generated in the first stage.
For example, Myers \etal \cite{myers} segmented a given food image, generated the food categories, the ingredients, and the corresponding volume.
The calories were then predicted by matching the estimated volume and the calorie density of the food category against a nutritional database.
Moreover, meta-data such as GPS location and user's food preferences were used to improve the predictions.
\looseness=-1


Although  multi-stage approaches improve several baseline methods for calorie estimation they have a number of disadvantages.
First, these methods require the definition and separate optimization at each stage, where each separate subtask in the different phases is difficult to solve accurately~\cite{subhi2019vision,min2019survey}.
Moreover, these models comprise a segmentation step for classifying the pixels into food and non-food categories and the estimation of the volume of the meal.
Thus, one would require pixel-wise annotations of a high amount of data as well as additional information specific to each meal in the image such as its volume.
Furthermore, multi-stage approaches require a definition of the different stages as well as the inputs and outputs for every stage, respectively.
This entails that potentially useful information from the earlier stages is not passed on and can not be used by the later stages to improve the predictions.\looseness=-1

\mypar{End-to-end Approaches for Food Related Tasks. }
End-to-end approaches rely on substituting the pipeline of multi-stage approaches by a single model.
Instead of training a pipeline of models addressing different subtasks, a single network is applied on the input data and \emph{directly} estimates the desired final output.
Thus, only the raw inputs and final outputs (next to the end-to-end architecture itself) need to be specified, while the neural network can learn which information is relevant internally.
End-to-end models do not struggle with the problems of multi-stage approaches, and have therefore replaced multi-stage approaches in a number of computer vision domains.
While end-to-end methods have shown great promise in tasks such as image recognition~\cite{resnet}, there are currently few end-to-end models for food related tasks such as calorie or nutrient prediction~\cite{situju2019food, takumi, chenDeepbasedIngredientRecognition2016}.

\mypar{Multi-task Approaches. }
Multi-task learning \cite{caruanaMultitaskLearning1997} is a method for addressing several different tasks concurrently with the overall aim of improving generalization by exploiting commonalities between separate problems.
\cite{chenDeepbasedIngredientRecognition2016,takumi} show that the performance of food related tasks such as ingredient prediction can be improved through multi-task learning.
Takumi \etal \cite{takumi} proposed a multi-task CNN which estimates the calories together with the average ingredient word embedding, the type of food, and an embedding of the preparation steps of the meal.
\cite{situju2019food} introduced a multi-task CNN to jointly predict the food category, calories, and the salinity given a food image which was evaluated on an unpublished dataset.
\looseness=-1 

\mypar{Recipe Datasets. }
Several works use variations of the Food-101 \cite{food101} dataset to address food image recognition tasks (\eg, \cite{myers,yanaiFoodImageRecognition2015,hassannejadFoodImageRecognition2016}).
The Food-101 dataset consists of altogether 101k pictures of dishes sorted into 101 categories.
Takumi \etal \cite{takumi} collected a dataset of English and Japanese recipes including ingredients and user-given calorie estimates that was not made publicly available.
Most other datasets consist of images captured in constrained environments, such as a specific canteen \cite{beijbom2015menu} or contain images of single food items instead of whole dishes \cite{liang2017computer}.
Furthermore, many datasets contain cuisine-specific data, such as images of just Chinese dishes \cite{chenDeepbasedIngredientRecognition2016} or American dishes \cite{chen2009pfid,food101}.
Hence, there is a lack of generic datasets containing various dishes from around the globe and recorded under unconstrained circumstances, \ie{} captured in the wild \cite{subhi2019vision}.
While most datasets contain food category and ingredient labels, there are few datasets containing nutritional and calorie annotations. \looseness=-1

The largest currently available dataset is Recipe1M \cite{marinRecipe1MDatasetLearning2019}, which encompasses one million images extracted from websites comprising categories, ingredients, and instructions.
The authors also expanded upon the dataset by adding nutritional information and pictures sourced from web searches using the recipe title (Recipe1M+).
Although this greatly increases the size of the dataset, the images are also far less relevant to the specific recipes and include images of similar dishes, prepackaged goods, and images taken by professional photographs.
The nutritional information was extracted using the USDA Food Database \cite{usda} by matching ingredient quantities and names semi-manually.
We compare our collected \dataset dataset to Recipe1M+ without web-search sourced pictures.
Within Recipe1M+, around 20k recipes with images have nutritional information, which totals to 50k images. 

\section{Dataset}
\label{sec:dataset}


\subsection{Data Collection}

We collect our dataset from a popular recipe website that contains ingredient lists, cooking instructions, and pictures of the resulting meals.
The recipes are from a large number of different cuisines, ranging from baked dishes to drinks, snacks, and dietary meals.
Most recipes have at least one picture, which are either uploaded by the original author of the recipe or by third parties following the instructions in the recipe.
Since the images are diverse and can comprise a single plate of food, while other consist of the entire dish, such as an entire casserole, it is not possible to directly obtain the information if an image contains a single portion or more.
Around $10$\% of recipes include a user-given value for the amount of calories in each portion that the recipe supposedly has.
We choose to use a German recipe website over American websites as a data source as American sites typically denote the required amount of a specific ingredient in  terms of its volume, \ie,  in number of cups or teaspoons, which are a more inaccurate measure with high variance in the actual mass.
Contrary, in German recipes, the amount of an item is mostly given by its mass, \ie{} in grams.
Our resulting dataset also covers  additional metadata  such  as  the  type of meal (\eg, dessert or side dish), the average user rating, the preparation time, and additional properties. \looseness=-1

\subsection{Preprocessing}
\label{sec:matching}

The dataset only has user-given calorie information for a small part of the data and does not include any details regarding the macronutrient composition.
Furthermore, since the information provided by the user is often inaccurate, we match the list of ingredients against a database of nutritional values to sum up the proportions of macronutrients as well as the total calories.
To facilitate this, we collect a secondary dataset from a German website of nutritional values, which contains values for the amount of fat, protein, and carbohydrates in grams per 100 grams of product.
The data is partially sourced from the USDA Food Composition Database \cite{usda}, and partially crowd sourced from manufacturer-given data of specific products.
Additionally, it contains food quantities such as ``1 medium-sized apple = 130g'', which we used to convert human-intuitive amounts that are commonly used in recipes (like \emph{add ``1 piece'' of fruit}) to their equivalent mass.
Thus, the processing of the recipes is comprised of two stages:

(1) Matching the ingredients to the nutritional database. 
For example, \emph{``medium-size white or Yukon Gold potatoes, peeled, cut into 1-inch cubes''} becomes \emph{potatoes}.

(2) Calculating the amount of different ingredients in the recipe. For example, \emph{``$1\sfrac{3}{4}$  cups diced potatoes''} is mapped to \emph{394 grams} in the second step of our proposed processing pipeline.
\looseness=-1

\mypar{Matching Ingredients. }
Since the ingredients are user-given and can be noisy, the mapping of the recipe ingredients to the nutritional database is not straightforward.
The given ingredient name often includes information that is not relevant to the product itself, but rather to its preparation or visual qualities.
These additional text snippets are hard to separate from information that is relevant, \eg, ``3 onions, diced'' and ``3 onions, in slices'' refer to the same product, while ``500g pasta, cooked'' and ``500g pasta, raw'' vary significantly in their calorie density.
Thus, since matching the textual representation of the ingredient  from a recipe to the ingredient from our nutritional database is ambiguous, we evaluate different matching methods on the text content of our dataset.

First, a simple option is to directly match the text of each ingredient  to the \emph{nearest} ingredient based on character edit distance.
However, this results in low quality matchings due to missing handling of synonyms and the above mentioned issue of irrelevant information in the description of the ingredient.
To address this, we tokenize the ingredient name to words and embed each word to a vector with Word2Vec~\cite{mikolov2013distributed} or FastText~\cite{fasttext}.
This is then averaged over the word vectors to get an ingredient embedding, in the same way as in the FastText library for extracting sentence vectors.
This still leads to unsatisfactory results, since each word in the ingredient name has the same weight, even though some words specify less important details.
For example in ``red onion'' vs ``red apple'', the word ``red'' is much less important than ``onion'' and ``apple''.
Finally, we obtain the best result by using the Google Universal Sentence Encoder \cite{googleuniv}, which can create $512$-dimensional embeddings for an arbitrary amount of input text.
We find the best matches for an ingredient by first comparing the embedding of the user-given free text from the recipe to the embeddings for all food items for which we have nutritional data using the cosine similarity (in our case this is equivalent to the dot product, since the embedded vectors are all normalized).
\looseness=-1

\mypar{Ingredient Amount Calculation.}
The second problem we have to tackle for pre-processing the dataset is calculating the normalized gram or milliliter amounts of ingredients in the recipes. 
For ingredients given in grams, this is trivial, but for many items the recipe authors use other units of measure, \eg, can, piece, tablespoon, some, or salt ``by taste''.
Since spices usually have little impact on the nutritional values, we exclude ingredients that are given ``by taste'' and similar.
For the other amounts, we match the unit name (like ``tablespoon'' or ``medium large'') exactly and multiply it with the given amount.
We also add some special cases like matching ``can'' to ``can (drained weight)'' and similar.
This amount matching is applied to all possible ingredients that have a cosine similarity of more than $0.84$ in descending order until a matching amount is found. 
If the amount matching fails, the ingredient is marked as unmatched, and if a recipe has at least one unmatched ingredient, it is completely discarded in our dataset.
We immediately discard recipes without perfect matches to be sure our dataset is accurate at the cost of losing a number of mostly good samples.
As a final step, we filter out all data points where the summed up calories of the recipe is outside of two standard deviations from the mean of all recipes.
We perform this step iteratively, until the outliers removal converges (\ie, no more outliers are found).
This is necessary because some recipes contain obviously wrong information (for example, in one carrot cake recipe the author specified to use a million carrots).
\looseness=-1

\subsection{Dataset Statistics}
\label{sec:stats}


\mypar{Overview. }
We found 211k recipes that contain at least a single image of the underlying recipe, of which around 20k recipes have user-given calorie information, which we, however, do not use for training our models.
In total, we collected 900k pictures where on average  each recipe has four images, while the recipes comprise a total of 374k \emph{unique} ingredients.
This high number is caused by slight differences in spelling or irrelevant details.
The database of nutritional values contains a total of 390k ingredients, which we filter by duplicates and popularity to 123k ingredients before matching the ingredients to the recipes.
After matching the ingredients found in the recipes with the food database, the count reduces to $5700$ unique ingredients.

\Cref{tbl:stats} shows statistics for each of the matching and preprocessing steps described in \cref{sec:matching}.
The final recipe count varies depending on whether we aggregate calories per recipe, per portion or per 100g of raw mass.
We discard around $60$\% of recipes during matching as we remove all recipes that have ingredients which do not fully match.
This is intentional as we want to ensure that we only retain data points that are accurate, and the proportion of usably matched recipes could be improved with further tweaking.
When aggregating per portion, we exclude more data points since we have to discard all recipes where the user did not supply information on how many portions a recipe contains.

Moreover, we show the amount of recipes discarded for each of our three proposed preprocessing steps.
First, 127k recipes contain an ingredient list that we were not able to completely match with the food database.
Even though imperfect matching of the ingredients and discounting unmatched ingredients would still lead to good estimates, we discard these cases to make sure that we do not worsen our approximation.

Finally, a total of 11k recipes do not contain any information per portion, so we cannot estimate the calorie values in these cases.
We experience strong outliers that are clearly incorrect in less than 21k recipes, which we discard in the final stage of our preprocessing pipeline. \looseness=-1

\mypar{Dataset Comparison. }
In total, our dataset contains 308k images with associated calorie estimate, as each of the 70k recipes includes on average multiple pictures.
\Cref{tbl:comparison} shows a comparison between our \dataset dataset and the Recipe1M+ dataset.
When excluding web-search sourced images, the Recipe1M+ dataset has around 20k recipes with images and nutritional information.
After applying the same outlier filtering to Recipe1M as for our own dataset, Recipe1M+ remains with 17k recipes with 44k images.
When aggregating per recipe, Recipe1M+ loses half the recipes (around 10k) during the outlier filtering.
The reason for this high number of excluded recipes is that the dataset includes a large amount of incorrectly parsed fractions (see \cite[sec. 4.1]{liDeepCookingPredicting2019}), which causes the ingredient amounts and thus the total mass to diverge by orders of magnitude in around $64$\%\footnote{We manually compared parsed ingredients of 100 randomly selected recipes of the Recipe1M+ dataset to their original on the websites.} of the parsed recipes (for example, \emph{$\sfrac{1}{4}$ cup} is mapped to \emph{14 cups}). \vspace{-1mm}


\begin{table}
\begin{center}
\vspace{0.17cm}
\caption{Preprocessing steps (each described in \cref{sec:matching}) and corresponding sample counts depending on the amount of food for that the calories are estimated. \label{tbl:stats}}
\vspace{-0.15cm}
\scalebox{1.15}{
\begin{tabular}{@{}p{2.74cm} ccc@{}}
\toprule
       Preprocessing                    & \!\!Per portion & Per 100 g & Per recipe \\
\midrule
Original recipes count       & 211k        & 211k      & 211k       \\
Removed (incomplete \newline ingredients match)      & \multirow{2}{*}{127k}        & \multirow{2}{*}{127k}      & \multirow{2}{*}{127k}       \\
Removed (no portion size information)        & \multirow{2}{*}{31k}         & \multirow{2}{*}{0}        & \multirow{2}{*}{0}         \\
Removed (kcal outliers)              & 11k         & 14k       & 21k        \\
\midrule
Final recipe count        & \textbf{42k}         & \textbf{70k}       & \textbf{63k}        \\
\bottomrule
\end{tabular}
}
\end{center}

\end{table}

\begin{table}
\caption{Comparison of sample and recipe counts of Recipe1M+ \cite{marinRecipe1MDatasetLearning2019} with our dataset. Recipe1M+ does not include portion information. Note that due to the parsing errors described in \cref{sec:stats} a large fraction of Recipe1M+ recipes included in these counts is not usable for training models relying on the ingredient amounts (such as calorie prediction).\label{tbl:comparison}}
\vspace{-0.25cm}
\begin{center}
\scalebox{1.15}{
\begin{tabular}{@{}l l c@{\hspace{2mm}}c@{\hspace{2mm}}c@{}}
    \toprule
         Dataset                 & Property& Per\! portion & Per\! 100g & Per recipe \\
\midrule

\multirow{4}{*}{Recipe1M+}\!\! & Mean [kcal]              & N/A & 219 & 1047  \\
 &Std. Dev. [kcal]\!\!\!           & N/A & 129 & 658  \\
&Recipe count & N/A & 17k & 10k\\
&Sample count & N/A & \textbf{44k} & \textbf{24k} \\
\midrule
\multirow{4}{*}{\dataset} &Mean [kcal]                   & 425     & 179  & 1791  \\
 &Std. Dev. [kcal]\!\!\!                & 207     & 73    & 1007  \\
&Recipe count        & 42k         & 70k       & 63k        \\
&Sample count & \textbf{179k}        & \textbf{308k}      & \textbf{267k}  \\
\bottomrule
\end{tabular}
}
\end{center}
\vspace{-0.6cm}

\end{table}


\mypar{Data Split. }
We split the recipes into train, validation, and test set  ($70$\%, $15$\%, and $15$\%, respectively), where we made sure that pictures of the same recipe are in the same data split. \vspace{-1mm}

\mypar{Ingredient Distribution. }
\Cref{tbl:ings} shows the frequency of the $15$ most frequent ingredients in our dataset.
As can be seen, spices (\eg, salt, pepper) and popular baking components (\eg, flour) are the strongest represented ingredients in our data. \looseness=-1


\begin{figure}
    \hypertarget{fig:model}{%
    \centering
    \includegraphics[width=0.8\linewidth]{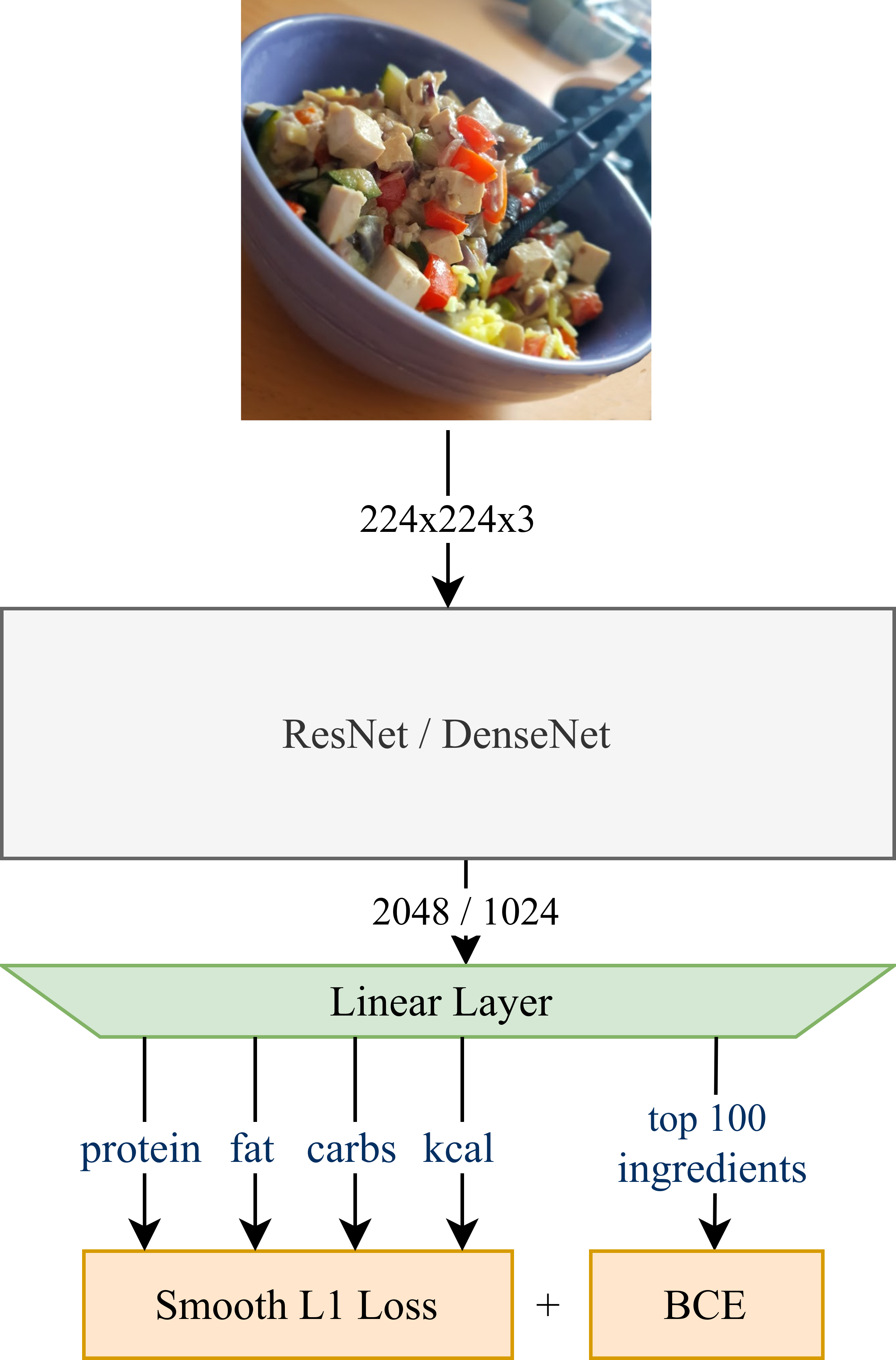}
    \caption{Overview of our proposed architecture. Our model is trained in a multi-task setup inferring the calories, the ingredients, and the macronutrients (e.g. protein).}\label{fig:model}
    }
    \vspace{-0.3cm}
\end{figure}

\section{Methods}
\label{sec:models}

In this section, we describe the architectures we adopt for calorie estimation as well as our enhancement of these models with the \emph{multi-task paradigm}.
We use end-to-end algorithms, \ie~the networks operate on the raw food  image without intermediate processing steps.
Similar to other multi-task approaches in food image processing \cite{chenDeepbasedIngredientRecognition2016, takumi, situju2019food}, we leverage pre-trained networks that are popular in the computer vision community.
For this, we adopt multiple variants of the prominent ResNet~\cite{resnet} and DenseNet~\cite{DenseNet} architectures as our backbone models, which were pre-trained on ImageNet~\cite{imagenet}.
As the original models were developed for \emph{classification}, while we aim for calorie \emph{regression}, we replace the last fully-connected layer with a regression layer and keep the rest of the architecture unchanged.
\looseness=-1

We argue that ground truth values of macronutrients correlate with the calorie amount and carry useful information for calorie estimation.
We therefore propose to include this additional knowledge as a supplemental learning signal to the calorie estimation network by following a \emph{multi-task learning paradigm}.
Our extension modifies the last fully connected layer of the network and does not alter the core architecture.
Originally, the models predict the calorie value (in kcal) by using a \emph{single output neuron}.
We enhance this architecture with \textit{three additional regression neurons for protein, fat, and carbohydrates} (in grams) in a multi-task network.

Our multi-task model regresses the macronutrient information with additional binary (sigmoid) outputs to predict the top $n=100$ ingredients.
The resulting final layer therefore has four regression outputs for the calories and macronutrients and $n$ binary outputs for the top $n$ ingredients (as we show in \cref{fig:model}).
We then compute the final loss by combining the smooth L1 loss for the regression outputs and the Binary Cross Entropy (BCE) loss for the top-100 ingredients:
\[ \text{multi-task loss} = \text{L1}_\text{kcal} + \sum_{m \in \{\text{fat},\text{prot},\text{carb}\}}{\text{L1}_m} + \gamma \cdot \text{BCE} \]

To get the same scaling of the learning signals, we multiply the BCE loss with a factor $\gamma$ depending on the dataset, therefore balancing the losses to have roughly equal contribution.

As there is no related work on published datasets targeting the same task with similar data, and no common baseline used in related work, we implement a \emph{baseline predicting the mean of all training samples} for comparison. This is a simple method that still includes the general statistical distribution of the
dataset. We also add an alternative baseline using the values from a random recipe in the dataset as the prediction. Since hitting the correct recipe is very unlikely, this baseline is equivalent to predicting a random gaussian value with the mean and variance of the dataset, which has worse performance than always predicting the mean value.
We use the same baselines for the macronutrient regression.
\looseness=-1

\begin{table}
    \caption{Most common ingredients after matching.\label{tbl:ings}}
    \vspace{-0.2cm}
    \begin{center}
    \scalebox{1.15}{
    \begin{tabular}{@{}l c@{}}
    \toprule
     Ingredient & Number of recipes\\
    \midrule
    Salt & 119244 \\
    Sugar & 59066 \\
    Chicken Egg &  58185\\
    Flour  & 46069\\
    Butter &  45891\\
    Onion, fresh  & 41206\\
    Milk  & 24531\\
    Vanilla Sugar  & 24011\\
    Oil  & 22822\\
    Paprika  & 22781\\
    Garlic  & 21348\\
    Water  & 20359 \\
    Pepper  & 19336\\
    Olive Oil  & 18928\\
    Baking Powder~~~~~~~~~~~  & 15966\\
    Cream  & 15039 \\
    \bottomrule
    \end{tabular}
    }
    \end{center}
    \vspace{-0.5cm}
\end{table}

\section{Evaluation}
\label{sec:eval}

\subsection{Setup}

To evaluate the model performance and create a competitive benchmark, we conduct extensive experiments with multiple neural networks (various ResNet~\cite{resnet} and DenseNet~\cite{DenseNet} variants). 
All models are implemented using PyTorch \cite{pytorch} and are trained for 25 epochs using a batch size of 50. 

Our initial goal was to predict the calories of a meal shown in an image. 
Since the dataset  does not provide information for how much food is in one specific image - and images can contain both a single served plate of food as well as a full pot - we run the following types of experiments:

\begin{itemize}
\item
    calorie prediction per recipe
\item
    calorie prediction per portion
\item
    calorie prediction per 100g of raw ingredient mass
\end{itemize}

Our first set of experiments leverages raw unfiltered data and results in an insufficient  performance of the models.  
Recognition results were further improved by adding the quality and outlier filtering described in \cref{sec:dataset}.
We also evaluated a model in a  classification setup instead of using the regression losses, however, the performance did not improve.


Furthermore, we run experiments to analyze the impact of additional information on the performance of our neural architectures. 
Within these experiments, we leverage the following three levels of additional information:

\begin{itemize}
\item
    \textbf{kcal-only:} As a reference, we predict the calories directly with no further information.
\item
    \textbf{kcal + nutritional information:} In addition to calories, the model predicts protein, fat, and carbohydrate amounts, which allows it to learn explicitly about the different components adding up to the food energy of a meal.
\item
    \textbf{kcal + nutritional information + top-100 ingredients:} The model also predicts which of the top-100 most common ingredients of the dataset are present in the dish. This further supports it in predicting the calories since it can more easily learn the correlation between different ingredients and the macronutrients.
\end{itemize}


\subsection{Results}

To evaluate the quality of  calorie prediction, we adopt the relative error ($\text{rel\_error} = 1 - |\frac{\text{pred}}{\text{truth}}|$) as our main metric.
In addition, we provide the absolute error (L1 error) of calories (in kcal), fat, protein, and carbohydrates (each in grams). 
In this setting, using the relative error is not very helpful, since in many dishes there is at least one absent macronutrient value.
In tables \ref{tbl:resbymodel},\ref{tbl:resperper}, and \ref{tbl:resbytask}, the first column depicts the \emph{relative} and the other columns show the \emph{absolute} error (either in kcal or grams). 
We compute the baselines as described in \cref{sec:models} and subsequently  compare the results for three different tasks. 

In table \ref{tbl:resbymodel}, we show the results of our baseline methods for the macronutrient and calorie estimation.
As expected, the random baseline has a poorer performance than the mean estimate having a discrepancy in absolute error of around $23$.
Then, we compare different backbone architectures for our multi-task setup. 
As we see, the two DenseNet~\cite{DenseNet} versions yield the highest performance (a relative error of $0.326$ and absolute error of $46.9$ kcal for calorie estimation in case of DenseNet121).
In contrast, the ResNet and ResNeXt \cite{resnext} architectures show an error of more than $0.33$.

Next, we compare  calorie regression for different amounts of food (\cref{tbl:resperper}), as the actual food quantity present in the photograph is not explicitly annotated in our dataset, and is only implicitly inferred via the portion- and recipe sizes.
The best performance is reached when predicting the amount of calories per 100g. 
This can be attributed to the following: 
(1) we collected the largest amount of data samples for the 100g setting.
(2) the size of a portion and of a recipe can be very subjective and noisy, while 100g is a more exact measurement.
Note that this remark also holds when using the baseline models (\cref{tbl:resperper}).

\begin{table}
\begin{center}
\caption{Relative and absolute regression error (in kcal / grams) by model architecture, each predicting calories, macronutrients, and ingredients per 100g in a multi-task setting. In this setup, DenseNet achieves the best performance. \label{tbl:resbymodel}}
\vspace{0.1cm}

\scalebox{1.1}{
\begin{tabular}{@{}l ccccc@{}}
\toprule
Model &  kcal (rel) &  kcal &  protein &  fat &  carbs \\
\midrule
Rand. Baseline & 0.595 & 83.3 & 4.36 & 6.32 & 15.0 \\
Mean Baseline & 0.464 & 60.5 & 3.10 & 4.49 & 10.5 \\
\midrule
ResNet50        &               0.334 &         47.8 &            2.54 &        3.93 &                  7.13 \\
ResNet101       &               0.336 &         48.2 &            2.54 &        3.94 &                  7.17 \\
ResNext50\_32$\times$4d\!\! &                0.33 &         47.2 &             2.50 &        3.89 &                  6.99 \\
\textbf{DenseNet121}      &               \textbf{0.326} &         46.9 &            2.51 &        3.88 &                  6.97 \\
DenseNet201     &               0.327 &         47.2 &            2.53 &        3.89 &                  7.04 \\
\bottomrule
\end{tabular}}
\end{center}


\begin{center}
\caption{Relative and absolute regression error (in kcal / grams) depending on the amount of food for which the nutritional values were predicted. Each model is predicting calories, macronutrients, and ingredients using a multi-task DenseNet121 backbone model. \label{tbl:resperper}}
\vspace{0.1cm}
\scalebox{1.1}{
\begin{tabular}{@{}l l ccccc@{}}

\toprule
{} & amount\!\! & \!kcal (rel)\! &  kcal &  protein &  fat &  carbs \\
\midrule

Rand. BL & \multirow{3}{*}{portion} & 0.909 & 235 & 15.5 & 16.0 & 30.8 \\
Mean BL & & 0.736 & 170 & 11.2 & 11.4 & 22.2 \\
\textbf{Ours}  &&              \textbf{0.623} &          154 &            9.21 &        10.7 &                  19.1 \\
\midrule
Rand. BL & \multirow{3}{*}{recipe} & 1.37 & 1130 & 57.7 & 73.0 & 167 \\
Mean BL & & 1.23 & 858 & 41.9 & 54.4 & 125 \\
\textbf{Ours}   &  &  \textbf{0.823} &          711 &            34.8 &        46.9 &                  94.4 \\
\midrule
Rand. BL & \multirow{3}{*}{100g} & 0.595 & 83.3 & 4.36 & 6.32 & 15.0 \\
Mean BL &  & 0.464 & 60.5 & 3.10 & 4.49 & 10.5 \\
\textbf{Ours}     &  &               \textbf{0.326} &         46.9 &            2.51 &        3.88 &                  6.97 \\
\bottomrule
\end{tabular}
}
\end{center}


\begin{center}

\caption{Relative and absolute regression error (in kcal / grams) by prediction task, each predicting values per 100g using a DenseNet121 backbone. The multi-task model that predicts the calories, macronutrients, and top-100 ingredients simultaneously has the best performance for all properties.\label{tbl:resbytask}}
\vspace{0.1cm}
\scalebox{1.1}{
\begin{tabular}{@{}l ccccc@{}}
\toprule
{} &  kcal (rel) &  kcal &  protein &  fat &  carbs \\
\midrule
Random Baseline~~~ & 0.595 & 83.3 & 4.36 & 6.32 & 15.0 \\
Mean Baseline & 0.464 & 60.5 & 3.10 & 4.49 & 10.5 \\
Kcal-only      &               0.362 &         50.3 &             N/A &         N/A &                   N/A \\
\"+ macros        &               0.345 &           49.0 &            2.67 &        4.06 &                   7.70 \\
\textbf{\"+\"+ top-100 ingredients}\!\!\!\!\!\!\!\!\!     &               \textbf{0.326} &         46.9 &            2.51 &        3.88 &                  6.97 \\
\bottomrule
\end{tabular}
}
\end{center}

\vspace{-0.4cm}
\end{table}

\begin{figure}  
    \sffamily\robotocondensed\small
\begin{tabular}{cc}
        \includegraphics[align=c,width=0.5\linewidth]{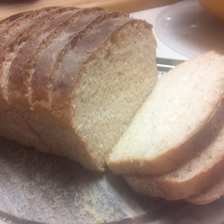} &    
    \vspace{-0.2cm}
        \begin{tabularx}{0.41\linewidth}{p{0.1\linewidth}>{\raggedleft}p{0.1\linewidth}>{\raggedleft\arraybackslash}p{0.1\linewidth}}
        \multicolumn{3}{l}{Nutrition Facts (per 100 g)} \\
        \toprule
        & Pred & True \\
        \midrule
        Calories & 229 kcal & 239 kcal \\
        \midrule
        Fat & 3 g & 2 g \\
        \midrule
        Carb & 44 g & 46 g \\
        \midrule
        Protein & 7 g & 7 g \\
        \midrule
        \multicolumn{3}{p{0.37\linewidth}}{\raggedright Ingredients (pred): Flour} \\
        \midrule
        \multicolumn{3}{p{0.37\linewidth}}{\raggedright Ingredients (true): Oil, Flour} \\
        \bottomrule
        \end{tabularx} \\[17ex]

    \vspace{-0.2cm}
        \includegraphics[align=c,width=0.5\linewidth]{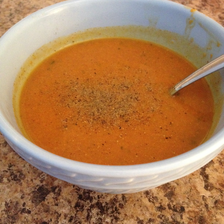} &    
        \begin{tabularx}{0.41\linewidth}{p{0.1\linewidth}>{\raggedleft}p{0.1\linewidth}>{\raggedleft\arraybackslash}p{0.1\linewidth}}
        \multicolumn{3}{l}{Nutrition Facts (per 100 g)} \\
        \toprule
        & Pred & True \\
        \midrule
        Calories & 99 kcal & 59 kcal \\
        \midrule
        Fat & 8 g & 4 g \\
        \midrule
        Carb & 7 g & 5 g \\
        \midrule
        Protein & 3 g & 1 g \\
        \midrule
        \multicolumn{3}{p{0.37\linewidth}}{\raggedright Ingredients (pred): } \\
        \midrule
        \multicolumn{3}{p{0.37\linewidth}}{\raggedright Ingredients (true): Garlic} \\
        \bottomrule
        \end{tabularx} \\[17ex]

    \vspace{-0.2cm}
        \includegraphics[align=c,width=0.5\linewidth]{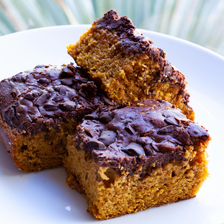} &    
        \begin{tabularx}{0.41\linewidth}{p{0.1\linewidth}>{\raggedleft}p{0.1\linewidth}>{\raggedleft\arraybackslash}p{0.1\linewidth}}
        \multicolumn{3}{l}{Nutrition Facts (per 100 g)} \\
        \toprule
        & Pred & True \\
        \midrule
        Calories & 252 kcal & 241 kcal \\
        \midrule
        Fat & 12 g & 9 g \\
        \midrule
        Carb & 28 g & 33 g \\
        \midrule
        Protein & 8 g & 6 g \\
        \midrule
        \multicolumn{3}{p{0.37\linewidth}}{\raggedright Ingredients (pred): Egg, Flour} \\
        \midrule
        \multicolumn{3}{p{0.37\linewidth}}{\raggedright Ingredients (true): Egg, Flour, Baking powder, Cacao powder} \\
        \bottomrule
        \end{tabularx} \\[17ex]

        \includegraphics[align=c,width=0.5\linewidth]{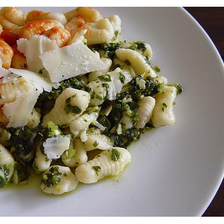} &    
        \begin{tabularx}{0.41\linewidth}{p{0.1\linewidth}>{\raggedleft}p{0.1\linewidth}>{\raggedleft\arraybackslash}p{0.1\linewidth}}
        \multicolumn{3}{l}{Nutrition Facts (per 100 g)} \\
        \toprule
        & Pred & True \\
        \midrule
        Calories & 190 kcal & 229 kcal \\
        \midrule
        Fat & 9 g & 17 g \\
        \midrule
        Carb & 20 g & 13 g \\
        \midrule
        Protein & 6 g & 4 g \\
        \midrule
        \multicolumn{3}{p{0.37\linewidth}}{\raggedright Ingredients (pred): } \\
        \midrule
        \multicolumn{3}{p{0.37\linewidth}}{\raggedright Ingredients (true): Onions, Garlic, Parsley} \\
        \bottomrule
        \end{tabularx} 

\end{tabular}

    \vspace{0.15cm}
    \caption{
        Example predictions of our multi-task model for different types of dishes. We report the predicted calories, fat, protein,
        carbohydrates, and ingredients from the validation set. 
        We show that the predictions are close to the ground truth, in particular the model can easily distinguish high- and low-calorie foods (\eg, between bread and soup). \looseness=-1
    }
    \label{fig:results}
\end{figure}

Finally, we examine the effect of the multi-task learning paradigm which we propose in \cref{sec:models}, \ie{} simultaneously predicting the macronutrient amounts (fat, carbohydrates, protein) and the presence of  ingredients in addition to the calorie amount (\cref{tbl:resbytask}). 
Multi-task learning clearly elevates the performance of our model in all prediction tasks.
For example, our multi-task model decreases the relative error by $9.9\%$, compared to the calorie-only approach. 
 This shows that the model learns the interplay between the amount of different macronutrients and the calorie quantity, as well as between the ingredients and the calories, even though how these are correlated (e.g. 1g of protein = 4kcal) is not explicitly defined during training.
\looseness=-1

In \cref{fig:results} we provide multiple examples of our model predictions for different food photographs and the corresponding ground truth. 
While our model does not give perfect calorie predictions, it discriminates well between high- and low-calorie dishes. 
For example, the soup image is assigned $99$ calories, contrary to a chocolate cake with a prediction of $252$ calories (the ground truth values are $59$ and $241$ kcal, respectively). 
In the examples, the model shows a better performance at predicting calories for baked goods, but has problems with dishes that have a greater variety of components, such as soups, where our model has not inferred any of the top-100 ingredients. This trend is reflected in the results on the whole dataset,
which is to be expected as baked dishes often require similar ingredients such as flour or eggs.
\looseness=-1




\section{Conclusion and Future Work}

\label{sec:conclusion}
We address the problem of end-to-end calorie estimation from food images. 
To that end, we propose a collection scheme for generating ground truth calorie estimates by matching the recipes to a food database.
We employ our collection scheme and create the large-scale \dataset dataset totaling 70\,000 recipes with 308\,000 associated images with ingredients and nutritional information.
We use the collected dataset to evaluate baselines and several neural networks previously designed for image classification.
We show that performance is improved through the use of multi-task learning, where the model simultaneously predicts the ingredients of the meal and macronutrients in addition to the calorie estimation.

In this work, even though we focus on calorie estimation, the dataset contains additional attributes that can characterize the meals.
In addition to calories, we include the type of the meal (such as dessert or side dish), ingredient quantities, cooking instructions, an average user rating, preparation time, tags, and further properties. 
Thus, a possible future work would be to use these additional properties to further improve the calorie prediction models in a similar manner to our multi-task setup where we leverage the macronutrients and ingredients.

The quality of the ingredient matching (\ref{sec:matching}) could be improved by fine-tuning the sentence embedding model used to match the free-form ingredient names by using the cooking instructions as a text corpus.
Further problems related to food computing could also be approached using the \dataset dataset, like detecting if a meal is vegan or fits a specific diet (low-carb, paleo, keto, etc.). 
Finally, we hope that the \dataset dataset paves the way for mobile nutrition applications as the data is collected in the wild, and most images are essentially captured with smartphone cameras.\looseness=-1


\balance
\bibliography{mybib}{}

\begin{thebibliography}{10}

\bibitem{beijbom2015menu}
Oscar Beijbom, Neel Joshi, Dan Morris, Scott Saponas, and Siddharth Khullar.
\newblock Menu-match: Restaurant-specific food logging from images.
\newblock In {\em 2015 IEEE Winter Conference on Applications of Computer
  Vision}, pages 844--851. IEEE, 2015.

\bibitem{fasttext}
Piotr Bojanowski, Edouard Grave, Armand Joulin, and Tomas Mikolov.
\newblock Enriching {Word} {Vectors} with {Subword} {Information}.
\newblock {\em Transactions of the Association for Computational Linguistics},
  5:135--146, December 2017.

\bibitem{food101}
Lukas Bossard, Matthieu Guillaumin, and Luc Van~Gool.
\newblock Food-101 -- mining discriminative components with random forests.
\newblock In {\em European Conference on Computer Vision}, 2014.

\bibitem{caruanaMultitaskLearning1997}
Rich Caruana.
\newblock Multitask learning.
\newblock {\em Machine learning}, 28(1):41--75, 1997.

\bibitem{googleuniv}
Daniel Cer, Yinfei Yang, Sheng-yi Kong, Nan Hua, Nicole Lyn~Untalan Limtiaco,
  Rhomni~St John, Noah Constant, Mario Guajardo-Céspedes, Steve Yuan, Chris
  Tar, Yun-hsuan Sung, Brian Strope, and Ray Kurzweil.
\newblock Universal {Sentence} {Encoder}.
\newblock In {\em In submission to: {EMNLP} demonstration}, Brussels, Belgium,
  2018.

\bibitem{chenDeepbasedIngredientRecognition2016}
Jingjing Chen and Chong-wah Ngo.
\newblock Deep-based {{Ingredient Recognition}} for {{Cooking Recipe
  Retrieval}}.
\newblock In {\em Proceedings of the 2016 {{ACM}} on {{Multimedia Conference}}
  - {{MM}} '16}, pages 32--41. {ACM Press}.

\bibitem{chen2009pfid}
Mei Chen, Kapil Dhingra, Wen Wu, Lei Yang, Rahul Sukthankar, and Jie Yang.
\newblock Pfid: Pittsburgh fast-food image dataset.
\newblock In {\em 2009 16th IEEE International Conference on Image Processing
  (ICIP)}, pages 289--292. IEEE, 2009.

\bibitem{chokr}
Manal Chokr and Shady Elbassuoni.
\newblock Calories {Prediction} from {Food} {Images}.
\newblock In {\em Proceedings of the {Thirty}-{First} {AAAI} {Conference} on
  {Artificial} {Intelligence}}, {AAAI}'17, pages 4664--4669, San Francisco,
  California, USA, 2017. AAAI Press.

\bibitem{takumi}
Takumi Ege and Keiji Yanai.
\newblock Image-{Based} {Food} {Calorie} {Estimation} {Using} {Knowledge} on
  {Food} {Categories}, {Ingredients} and {Cooking} {Directions}.
\newblock In {\em Proceedings of the on {Thematic} {Workshops} of {ACM}
  {Multimedia} 2017}, Thematic {Workshops} '17, pages 367--375, New York, NY,
  USA, 2017. ACM.

\bibitem{hassannejadFoodImageRecognition2016}
Hamid Hassannejad, Guido Matrella, Paolo Ciampolini, Ilaria De~Munari, Monica
  Mordonini, and Stefano Cagnoni.
\newblock Food {{Image Recognition Using Very Deep Convolutional Networks}}.
\newblock In {\em Proceedings of the 2nd {{International Workshop}} on
  {{Multimedia Assisted Dietary Management}}}, {{MADiMa}} '16, pages 41--49.
  {Association for Computing Machinery}.

\bibitem{resnet}
K.~He, X.~Zhang, S.~Ren, and J.~Sun.
\newblock Deep {Residual} {Learning} for {Image} {Recognition}.
\newblock In {\em 2016 {IEEE} {Conference} on {Computer} {Vision} and {Pattern}
  {Recognition} ({CVPR})}, pages 770--778, June 2016.

\bibitem{resnext}
Kaiming He, Zhuowen Tu, Piotr Dollár, Ross Girshick, and Saining Xie.
\newblock Aggregated {Residual} {Transformations} for {Deep} {Neural}
  {Networks}.
\newblock November 2016.

\bibitem{DenseNet}
G.~Huang, Z.~Liu, L.~v~d Maaten, and K.~Q. Weinberger.
\newblock Densely {Connected} {Convolutional} {Networks}.
\newblock In {\em 2017 {IEEE} {Conference} on {Computer} {Vision} and {Pattern}
  {Recognition} ({CVPR})}, pages 2261--2269, July 2017.

\bibitem{noauthor_calorie_nodate}
2017~Azumio Inc.
\newblock Calorie {Mama} {Food} {AI} - {Food} {Image} {Recognition} and
  {Calorie} {Counter} using {Deep} {Learning}.
\newblock \url{https://caloriemama.ai/}, [Accessed: 2019-08-17].

\bibitem{liDeepCookingPredicting2019}
Jiatong Li, Ricardo Guerrero, and Vladimir Pavlovic.
\newblock Deep {{Cooking}}: {{Predicting Relative Food Ingredient Amounts}}
  from {{Images}}.
\newblock In {\em Proceedings of the 5th {{International Workshop}} on
  {{Multimedia Assisted Dietary Management}}}, {{MADiMa}} '19, pages 2--6.
  {Association for Computing Machinery}.

\bibitem{liang2017computer}
Yanchao Liang and Jianhua Li.
\newblock Computer vision-based food calorie estimation: dataset, method, and
  experiment.
\newblock {\em arXiv preprint arXiv:1705.07632}, 2017.

\bibitem{marinRecipe1MDatasetLearning2019}
Javier Marin, Aritro Biswas, Ferda Ofli, Nicholas Hynes, Amaia Salvador, Yusuf
  Aytar, Ingmar Weber, and Antonio Torralba.
\newblock {{Recipe1M}}+: {{A Dataset}} for {{Learning Cross}}-{{Modal
  Embeddings}} for {{Cooking Recipes}} and {{Food Images}}.
\newblock pages 1--1.

\bibitem{mikolov2013distributed}
Tomas Mikolov, Ilya Sutskever, Kai Chen, Greg~S Corrado, and Jeff Dean.
\newblock Distributed representations of words and phrases and their
  compositionality.
\newblock In {\em Advances in neural information processing systems}, pages
  3111--3119, 2013.

\bibitem{min2019survey}
Weiqing Min, Shuqiang Jiang, Linhu Liu, Yong Rui, and Ramesh Jain.
\newblock A survey on food computing.
\newblock {\em ACM Computing Surveys (CSUR)}, 52(5):1--36, 2019.

\bibitem{myers}
A.~Myers, N.~Johnston, V.~Rathod, A.~Korattikara, A.~Gorban, N.~Silberman,
  S.~Guadarrama, G.~Papandreou, J.~Huang, and K.~Murphy.
\newblock Im2calories: {Towards} an {Automated} {Mobile} {Vision} {Food}
  {Diary}.
\newblock In {\em 2015 {IEEE} {International} {Conference} on {Computer}
  {Vision} ({ICCV})}, pages 1233--1241, December 2015.

\bibitem{usda}
US~Department of~Agriculture, Agricultural~Research Service, and Nutrient~Data
  Laboratory.
\newblock {USDA} {Food} {Composition} {Databases}.
\newblock \url{https://ndb.nal.usda.gov/ndb/}, [Accessed: 2019-08-04].

\bibitem{pytorch}
Adam Paszke, Sam Gross, Francisco Massa, Adam Lerer, James Bradbury, Gregory
  Chanan, Trevor Killeen, Zeming Lin, Natalia Gimelshein, Luca Antiga, Alban
  Desmaison, Andreas Kopf, Edward Yang, Zachary DeVito, Martin Raison, Alykhan
  Tejani, Sasank Chilamkurthy, Benoit Steiner, Lu~Fang, Junjie Bai, and Soumith
  Chintala.
\newblock Pytorch: An imperative style, high-performance deep learning library.
\newblock In H.~Wallach, H.~Larochelle, A.~Beygelzimer, F.~d\textquotesingle
  Alch\'{e}-Buc, E.~Fox, and R.~Garnett, editors, {\em Advances in Neural
  Information Processing Systems 32}, pages 8024--8035. Curran Associates,
  Inc., 2019.

\bibitem{pouladzadehUsingGraphCut2014}
Parisa Pouladzadeh, Shervin Shirmohammadi, and Abdulsalam Yassine.
\newblock Using graph cut segmentation for food calorie measurement.
\newblock In {\em 2014 {{IEEE International Symposium}} on {{Medical
  Measurements}} and {{Applications}} ({{MeMeA}})}, pages 1--6. {IEEE}.

\bibitem{imagenet}
Olga Russakovsky, Jia Deng, Hao Su, Jonathan Krause, Sanjeev Satheesh, Sean Ma,
  Zhiheng Huang, Andrej Karpathy, Aditya Khosla, Michael Bernstein,
  Alexander~C. Berg, and Li~Fei-Fei.
\newblock {ImageNet Large Scale Visual Recognition Challenge}.
\newblock {\em International Journal of Computer Vision (IJCV)}, 2015.

\bibitem{situju2019food}
Sulfayanti~F Situju, Hironori Takimoto, Suzuka Sato, Hitoshi Yamauchi, Akihiro
  Kanagawa, and Armin Lawi.
\newblock Food constituent estimation for lifestyle disease prevention by
  multi-task cnn.
\newblock {\em Applied Artificial Intelligence}, 33(8):732--746, 2019.

\bibitem{subhi2019vision}
Mohammed~Ahmed Subhi, Sawal~Hamid Ali, and Mohammed~Abulameer Mohammed.
\newblock Vision-based approaches for automatic food recognition and dietary
  assessment: A survey.
\newblock {\em IEEE Access}, 7:35370--35381, 2019.

\bibitem{yanaiFoodImageRecognition2015}
Keiji Yanai and Yoshiyuki Kawano.
\newblock Food image recognition using deep convolutional network with
  pre-training and fine-tuning.
\newblock In {\em 2015 {{IEEE International Conference}} on {{Multimedia}} \&
  {{Expo Workshops}} ({{ICMEW}})}, pages 1--6. {IEEE}.

\bibitem{zhou2018calorie}
Jun Zhou, Dane Bell, Sabrina Nusrat, Melanie Hingle, Mihai Surdeanu, and
  Stephen Kobourov.
\newblock Calorie estimation from pictures of food: Crowdsourcing study.
\newblock {\em Interactive journal of medical research}, 7(2):e17, 2018.

\bibitem{zhuIMAGEANALYSISSYSTEM2010}
Fengqing Zhu, Marc Bosch, Carol~J. Boushey, and Edward~J. Delp.
\newblock {{An immage analysis system for dietary assessment and evaluation}}.
\newblock pages 1853--1856.

\end{thebibliography}
\bibliographystyle{plain}
\end{document}